\documentclass{article}

\usepackage{PRIMEarxiv}

\usepackage{xcolor}
\usepackage[utf8]{inputenc} % allow utf-8 input
\usepackage[T1]{fontenc}    % use 8-bit T1 fonts
\usepackage{hyperref}       % hyperlinks
\usepackage{url}            % simple URL typesetting
\usepackage{booktabs}       % professional-quality tables
\usepackage{amsfonts}       % blackboard math symbols
\usepackage{nicefrac}       % compact symbols for 1/2, etc.
\usepackage{microtype}      % microtypography
\usepackage{fancyhdr}       % header
\usepackage{graphicx}       % graphics
\usepackage{xcolor}
\usepackage[numbers]{natbib}

\usepackage{booktabs}
\usepackage{cleveref}
\usepackage{ulem}

\graphicspath{{media/}}     % organize your images and other figures under media/ folder

\newcommand{\gptmodel}[0]{GPT-3.5}
\newcommand{\gptmodelrelease}[0]{GPT-3.5-turbo-0301}
\newcommand{\tf}[0]{$T$}
\newcommand{\es}[0]{$E$}
\newcommand{\ei}[0]{$e_{i}$}
\newcommand{\ej}[0]{$e_{j}$}
\newcommand{\te}[0]{$T^{G}_{i}$}
\newcommand{\esi}[0]{$E_{i}$}
\newcommand{\re}[0]{$R_{i}$}

\newcommand{\qt}[1]{``{#1}''}

\title{Iterative Zero-Shot LLM Prompting for Knowledge Graph Construction
}

\author{
  Salvatore Carta \\
  Department of Mathematics\\ 
  and Computer Science \\ 
  University of Cagliari \\
  Palazzo delle Scienze, Via Ospedale,\\ 
  72, 09124, Cagliari, Italy\\
  \texttt{salvatore@unica.it} \\
   \And
  Alessandro Giuliani,\\
  Department of Mathematics\\ 
  and Computer Science \\ 
  University of Cagliari \\
  Palazzo delle Scienze, Via Ospedale,\\ 
  72, 09124, Cagliari, Italy\\
  \texttt{alessandro.giuliani@unica.it} \\
   \And
  Leonardo Piano\\
  Department of Mathematics\\ 
  and Computer Science \\ 
  University of Cagliari \\
  Palazzo delle Scienze, Via Ospedale,\\ 
  72, 09124, Cagliari, Italy\\
  \texttt{leonardo.piano@unica.it} \\
   \And
  Alessandro Sebastian Podda \\
  Department of Mathematics\\ 
  and Computer Science \\ 
  University of Cagliari \\
  Palazzo delle Scienze, Via Ospedale,\\ 
  72, 09124, Cagliari, Italy\\
  \texttt{sebastianpodda@unica.it} \\
   \And
  Livio Pompianu\\
  Department of Mathematics\\ 
  and Computer Science \\ 
  University of Cagliari \\
  Palazzo delle Scienze, Via Ospedale,\\ 
  72, 09124, Cagliari, Italy\\
  \texttt{livio.pompianu@unica.it} \\
   \And
  Sandro Gabriele Tiddia \\
  Department of Mathematics\\ 
  and Computer Science \\ 
  University of Cagliari \\
  Palazzo delle Scienze, Via Ospedale,\\ 
  72, 09124, Cagliari, Italy\\
  \texttt{ sandrog.tiddia@unica.it} \\
}

\begin{document}
\maketitle

\begin{abstract}
In the current digitalization era, capturing and effectively representing knowledge is crucial in most real-world scenarios. In this context, knowledge graphs represent a potent tool for retrieving and organizing a vast amount of information in a properly interconnected and interpretable structure. However, their generation is still challenging and often requires considerable human effort and domain expertise, hampering the scalability and flexibility across different application fields. This paper proposes an innovative knowledge graph generation approach that leverages the potential of the latest generative large language models, such as GPT-3.5, that can address all the main critical issues in knowledge graph building. The approach is conveyed in a pipeline that comprises novel iterative zero-shot and external knowledge-agnostic strategies in the main stages of the generation process. Our unique manifold approach may encompass significant benefits to the scientific community. In particular, the main contribution can be summarized by: (i) an innovative strategy for iteratively prompting large language models to extract relevant components of the final graph; (ii) a zero-shot strategy for each prompt, meaning that there is no need for providing examples for ``guiding'' the prompt result; (iii) a scalable solution, as the adoption of LLMs avoids the need for any external resources or human expertise. To assess the effectiveness of our proposed model, we performed experiments on a dataset that covered a specific domain. We claim that our proposal is a suitable solution for scalable and versatile knowledge graph construction and may be applied to different and novel contexts.
\end{abstract}

% keywords can be removed

\section{Introduction}
\label{sec:introduction}
Nowadays, the world is experiencing an unprecedented transformation since the development of recent generative large language models (LLMs). The capability of such AI systems, like \gptmodel{}, in analyzing, processing, and comprehending a considerable amount of human language leads to rapid and strong improvements in several application fields. LLMs are changing the way of accessing, ingesting, and processing information. Indeed, they can enable humans to interact easily with machines, which can receive and understand the users' queries expressed in natural language and generate coherent, reliable, and contextually relevant responses represented in a proper natural language format.
The impact of LLMs transcends pure language processing. Indeed, they may lay the foundations for a not-distant future where the interaction and cooperation between humans and machines represent the primary strategy for innovation and progress. In this scenario, the potential of LLMs may revolutionize all real-world domains, e.g., technological applications, healthcare, or creative industries like music composition or even art.
Like classical Machine Learning algorithms, which are trained on large datasets to provide predictions given a specific input \cite{bib:sarker21}, LLMs are trained on vast written language datasets to predict natural language elements following a given input textual query (i.e., a \textit{prompt}), such as a question, instruction or statement. LLMs can generate a remarkable variety of outputs depending on the given prompt. Therefore, LLM prompting assumes a central role, and researchers are ever more focused on exploiting their potential to devise innovative algorithms and tools and improve knowledge in various domains. LLM prompting can be leveraged to enhance various Machine Learning tasks, as the principal strength is the ability to generate high-quality synthetic data reducing the manual effort in collecting and annotating data.
Indeed, by composing well-formulated prompts, LLMs can be highly valuable in supporting the devising of models in many scenarios, e.g., in generating creative content (like stories or poems) \cite{bib:peng22}, Information Retrieval (where users can request information on specific topics) \cite{bib:bonifacio20}, problem-solving \cite{bib:yao23}, or text summarization \cite{bib:zhang23}.

Notwithstanding the general encouraging performance of LLMs, their usefulness is still not optimal or under investigation in several areas. In this context, this paper aims to provide an innovative approach to knowledge representation, which plays an essential role in many real-world scenarios, transforming how the information is collected, organized, processed, and used. In particular, we focus on a novel method for creating suitable Knowledge Graphs (KGs), that have been demonstrated to be extremely valuable across several industries and domains. KGs organize information in a proper graph structure, where nodes represent entities and edges represent relations between entities. Each node and edge may be associated with properties or attributes, providing further details or metadata. KGs are powerful tools for capturing and representing knowledge appropriately that hold considerable advantages: 

\begin{itemize}
	\item they can infer and integrate information from heterogeneous sources, e.g., structured databases, unstructured text, or web resources;
	\item they can capture both explicit and implicit knowledge. Explicit information is directly represented in the KG (e.g., ``Joe Biden is the president of the USA''), whereas the implicit knowledge can be inferred by exploring the graph and finding relevant patterns and paths, allowing to discover new insights;
	\item they allow a straightforward and effective query and navigation of information. 
\end{itemize}

KGs are widely applied in many domains, e.g., in recommender systems \cite{bib:du23}, semantic search \cite{bib:wu23}, healthcare and clinical tools \cite{bib:abu-salih23}, or finance \cite{bib:hou22}.

Although significant improvements have been made in graph construction, creating a complete and comprehensive KG in an open-domain setting involves several challenges and limitations. For example, the lack of established large annotated datasets for relation extraction, which arises from the absence of a rigorous definition of what constitutes a valid open-domain relational tuple, leads to the development of relation extraction tools using heterogeneous datasets, which tend to work well on their trained dataset but are prone to produce incorrect results when used on different genres of text \cite{Niklaus2018ASO}. Moreover, the open-domain implementations of the complementary tasks of Named Entity Recognition and Entity Resolution are understandably less performing than their closed-domain counterparts, and they also suffer from the lack of well-established toolkits and comprehensive knowledge bases.

To overcome the aforementioned limitations, our insight is to rely on LLMs to support the construction of KGs. Indeed, their ability to analyze and generate human-like text at a large scale can be instrumental in knowledge representation. We deem that LLMs can enhance the KG generation in the main stages of the process, e.g., in extracting entities and relations, disambiguation, or textual inference.
In particular, relying on LLM prompting may be a key point of KG generation. By composing an appropriate prompt, LLMs can efficiently process structured and unstructured text and transform it into KG components. A well-formed prompt can lead to extract relevant entities, relationships, and types.

Summarizing, we present a novel approach for KG construction based on extracting knowledge with the support of an iterative zero-shot (i.e., without the need for any examples and fine-tuning) LLM prompting. In particular, a sequence of appropriate prompts is used iteratively on a given set of input documents to extract relevant triplets and their attributes for composing a KG. Subsequently, supported by a further prompting strategy, we defined a proper entity/predicate resolution method for resolving the entity/relations co-references. Finally, a different prompting approach is applied to develop an inference-based technique for defining a meaningful schema. To our knowledge, although using LLMs is a hot research topic, this is the first attempt at using them to develop a suitable approach for creating a complete KG.

The rest of the paper is organized as follows: Section~\ref{sec:rel_work} reports the  Knowledge Graph generation state-of-the-art. Section~\ref{sec:aims} summarizes our claim and research goals, whereas the proposed methodology is described in Section~\ref{sec:methodology}. Section~\ref{sec:experiments} is aimed at reporting all the experimental results and the related discussion. Section~\ref{sec:conclusion} ends the paper with the conclusions.

\section{Related Work}
\label{sec:rel_work}

Automatic knowledge graph construction aims to create a structured knowledge representation from different data sources without manual intervention.
Knowledge Graph Construction (KGC) pipelines typically employ Named Entity Recognition \cite{Chiu2016}, Relation Extraction \cite{Nguyen2015}, and Entity Resolution \cite{DeepERD} techniques, in order to transform unstructured text into a structured representation that captures the entities, their relationships, and associated attributes.
The pipeline of \cite{elhammadi2020high} combined co-reference resolution, named entities recognition, and Semantic Role Labeling to build a financial news Knowledge Graph.
\citeauthor{luan2018multi} \cite{luan2018multi} developed a multi-task model for identifying entities, relations, and coreference clusters in scientific articles able to support the creation of Scientific Knowledge Graphs. \citeauthor{mehta2019scalable} \cite{mehta2019scalable} presented an end-to-end KG construction system and a novel Deep Learning-based predicate mapping model. Their system identifies and extracts entities and relationships from text and maps them to the DBpedia namespace. Although these pipelines can achieve satisfactory results and produce high-quality Knowledge Graphs, their methods are often limited to a predefined set of entities and relationships or dependent on a specific ontology. Our proposal addresses these limitations, as we do not rely on predefined sets or external ontologies.
Additionally, almost all methodologies exploit a supervised approach, requiring extensive human manual annotation. To address this issue, since the introduction of the first Pre-Trained Language Models, the question has been:
\textit{can we use the knowledge stored in pre-trained LMs to construct KGs?} \citeauthor{wang2020language} have been among the first to raise and address such a question. They designed an unsupervised approach called \textit{MAMA} that constructs Knowledge Graphs with a single forward pass of the pre-trained LMs over the corpora without any fine-tuning \cite{wang2020language}.
\citeauthor{hao2022bertnet} \cite{hao2022bertnet} proposed a novel method for extracting vast Knowledge Graphs of arbitrary relations from any pre-trained LMS by using minimal user input. 
Given the minimal definition of the input relation as an initial prompt and some examples of entity pairs, their method generates a set of new prompts that can express the target relation in a diverse way.
The prompts are then weighted with confidence scores, and the LM is used to search a large collection of candidate entity pairs, followed by a ranking that yields the top entity pairs as the output knowledge.
Recent technological and scientific advancements accompanied by increasing data availability have led to a severe escalation in developing Large Language Models.
New LLMs such as GPT-3.5 have shown remarkable zero and few-shot Information Extraction capabilities, as demonstrated by \citeauthor{agrawal2022large} \cite{agrawal2022large} and \citeauthor{Wei2023ZeroShotIE} \cite{Wei2023ZeroShotIE}. \citeauthor{Li2023EvaluatingCI} \cite{Li2023EvaluatingCI} systemically assessed ChatGPT performance in various IE tasks. They pointed out that ChatGPT performs very well on Open Information Extraction and other simple IE tasks (e.g., entity typing) but struggles with more complex and challenging tasks such as Relation Extraction and Event Extraction. \citeauthor {Wan2023GPTREIL}have resolved this issue \cite{Wan2023GPTREIL}, identifying an in-Context Learning strategy to bridge the gap between LLMs and fully-supervised baselines reaching SOTA results in diverse literature RE datasets.
\citeauthor{ashok2023promptner} exploited LLM to provide an alternate way to approach the few-shot NER making it easily adjustable across multiple domains \cite{ashok2023promptner}. Their method, PromptNER, prompts an LLM to generate a comprehensive list of potential entities, accompanied by explanations that support their compatibility with the given definitions for entity types. Similarly, \citeauthor{ji2023vicunaner} introduced VicunaNER \cite{ji2023vicunaner}, a zero/few-shot NER framework based on the open-source LLM, Vicuna.
\citeauthor{trajanoska2023enhancing} proposed an improved KGC pipeline supported by LLM to organize and connect concepts \cite{trajanoska2023enhancing}, prompting ChatGPT to extract entities and relationships jointly and then preprocessing entities with an entity-linking strategy, where the entities having the same DBpedia URI are considered conceptually equal. The authors evaluated their system on a sustainability-related text use case. Nevertheless, their method has a crucial weakness, i.e., as it relies on an external knowledge base, many relevant entities not included in DBpedia are missed. Our approach, compliant with their method in some steps, aims to overcome such drawbacks.
Finally, \citeauthor{bi2023codekgc} proposed an innovative LLM-powered KGC method \cite{bi2023codekgc}. As LLMs have demonstrated excellent abilities in structure predictions, they leveraged GPT-3.5 code generation to create a Knowledge Graph by converting natural language into code formats. Converting natural language to code formats could make the model better capture structural and semantic information. Using structural code prompts and encoding schema information, they aim to enhance the modeling of predefined entity and relation schemas in knowledge graphs. In accordance with this work, our approach proposes an innovative LLM-based schema generation, as described in the following Sections.

\section{Research Aims and Motivations}
\label{sec:aims}
In this Section, we report the motivations and the research aims, starting by describing the open challenges in the scenario of KG generation.

\subsection{Problem statement}
\label{sec:problem}
The current solutions in constructing KGs, either domain-specific or general-purpose, address the task from many perspectives. This is also due to the need to discover the right tradeoff between providing high data quality, scalability, and automation \cite{bib:weikum21}. The variety of state-of-the-art approaches is a clear indicator of the intrinsic complexity of the problem. Therefore, many open challenges are currently leading to a demand for more research efforts. 
In particular, several factors affect the process of building a KG:

\begin{itemize}
	\item \textit{Data availabity and acquisition}. Accessing and adopting relevant and suitable data sources is still challenging. One of the main goals of a KG is to capture an exhaustive range of knowledge from various domains and sources, either structured or unstructured. However, the availability of data is typically hampered by several factors. For example, a common scenario is when data sources are restricted or inaccessible due to policies or privacy concerns, or data may be strewed across different repositories, platforms, formats, or even languages, making their integration very challenging. Furthermore, data is highly dynamic nowadays, with the risk of existing data becoming obsolete. Therefore, the current challenge is to put extensive effort into data acquisition, monitoring, and updating.
	\item \textit{Data quality}. The acquired data should be accurate, complete, consistent, and reliable to build and maintain a KG properly. Indeed, the quality of such data highly affects the effectiveness and reliability of KGs. Addressing data issues like incorrect or outdated information, insufficient or missing data, unreliable sources, or contradictory data is crucial in KG generation. To ensure richness and accuracy, a classical approach is to involve humans in annotating and labeling data. Leveraging human knowledge allows for a better understanding and context-specific interpretation of the input data. However, this process may not be affordable as it heavily depends on the availability of knowledgeable annotators and their time and effort. Therefore, guaranteeing reliable data and limiting human efforts, especially in the case of unstructured text data, is crucial.
	\item \textit{Scalability}. Providing and maintaining a high-quality KG, also in the case of automatic data acquisition and integration, is a complex goal due to the data dimensionality and heterogeneity of data sources. Indeed, scalability is crucial in a real-world scenario where an enormous amount of data must be analyzed. In general, the more the graph grows, the more the computation requires resources and time consumption.
	\item \textit{Subjectivity and contextual knowledge}. Although KGs mainly focus on factual information, representing subjective or context-dependent knowledge is extremely challenging and often requires further contextual information or external resources;
	\item \textit{Semantic disambiguation}. Natural language is intrinsically ambiguous. Semantic disambiguation, i.e., disambiguating words, resolving synonyms, handling polysemous terms, and capturing fine-grained unlikenesses in meaning, are current research challenges.
	\item \textit{Domain-specific expertise}. Different domains may retain complex and specialized knowledge structures, demanding domain expertise and more specialized strategies for knowledge inference. Domain experts may ensure quality and effectiveness but usually require a significant human effort, as already pointed out.
	%\item \textit{Incremental knowledge integration}. As already mentioned, due to the high dynamicity of data, even when generating a complete, reliable, and well-formed KG, it should be possible to integrate new knowledge in the final graph properly. Most current approaches do not support this task and rely on a ``batch-like'' regeneration of the KG.
        \item \textit{External resources}. The current methods of KG generation often rely on \textit{entity linking}, i.e., resolving entities and relations to external knowledge base (KB) concepts. The main weakness of such methods is the high risk of filtering out many relevant entities or concepts not included in the reference KB. On the other hand, a common way to extract triplets without relying on external knowledge is the adoption of Open Information Extraction (OpenIE), which discovers a wide range of triplets without prior knowledge. However, OpenIE methods usually generate a high number of incorrect or incomplete triplets; this behavior leads to including incorrect and misleading information in the KG.
	\item \textit{Evaluation}. Evaluating a KG is crucial to assess whether 
the generated KG properly represents the knowledge of the underlying scenario. However, evaluation is a challenging task due to several issues. First, there are often no ground truths or golden standards to adopt for comparisons. Furthermore, there are no standard metrics, and evaluating general-purpose methods across different use cases or application fields may be complex. 
	\item \textit{Pipeline definition}. While, typically, most parts of a specific KG generation pipeline are well-known methods, algorithms, or models, with exhaustive previous research, the interaction and the integration of all tasks is a current challenge.
\end{itemize}

\subsection{Research questions}
\label{sec:research_questions}
According to the aforementioned challenges and limitations, this work proposes an approach to address various issues in KG construction. In particular, we intend to answer to the following research question:

\begin{itemize}
	\item How can we effectively extract, analyze and enhance information from multiple textual data sources? 
	\item How can we improve the quality of the extracted information avoiding the need for human effort, also in case of generating KGs that usually require specific human expertise?
	\item How can we generate relevant and proper triplets without relying on external KBs or OpenIE methods?
	\item Regarding scalability, which strategies can we define for dealing with large-scale datasets for generating KGs with millions or even billions of entities and relationships?
	\item What methods can be devised to adequately perform disambiguation, entity resolution, and linking to properly represent the knowledge?
	%\item \warn{How can we fulfill the requirement of ``updating'' a generated KG with new information, with the goal of capturing temporal and contextual evolution of data and scenarios?}{\sgtcm{Questo è potenzialmente ok, ma va effettivamente raccontato, per ora non lo abbiamo fatto.}}
	\item How can we properly evaluate the generated KG also without priorly having a gold standard or a specific ground truth?%{\sgtcm{Non sono convito sia stata una domanda che ci ha guidato, dato che alla fine abbiamo valutato manualmente ogni componente generata una ad una, che non è nulla di innovativo o interessante.} (ALE: ma un gorund truth è stato comunque generato, e ci ha permesso di valutare la recall.) \sgtcm{Sì, ma la domanda dice proprio  \qt{\emph{without} a specific ground truth} e noi abbiammo proprio generato un ground truth.}}
\end{itemize}

\subsection{Main contribution}
\label{sec:contribution}
The previous research questions have guided the investigation of innovative methods for proposing proper strategies to address the main problems and open challenges. Let us note that, in this work, we focus on extracting information from heterogeneous unstructured textual documents. All our solutions have been incorporated into a novel pipeline for generating KGs, described in Section~\ref{sec:methodology}, that (i) relies on adopting an LLM that is iteratively queried with a sequence of adequately well-formed prompts to perform the main stages of the KG generation and (ii) can effectively perform an automated entity/predicate resolution without the need of external knowledge-bases or any human effort. To our knowledge, at the state-of-the-art, no other KG generation models rely on using our LLM prompting strategy in combination with an automated and knowledge-base agnostic entity resolution for a complete KG generation pipeline. 
Let us point out that in this preliminary work, we have assumed GPT-3.5 as LLM.

The main contribution of the paper is summarized in the following:

\begin{itemize}
	\item we propose an iterative LLM prompting-based pipeline for automatically generating KGs. Let us remark that there is no need for any human effort;
	\item we propose a sequence of proper, well-formed LLM prompts for each stage of the process. The devised prompts are able to:
		\begin{itemize} 
        	\item identify relevant entities and extract their descriptions and type;
			\item identify meaningful relationships and their descriptions;
			\item given the previous components, identify relevant triplets;
			\item provide domain-specific triplets, even if no information about the domain is given as input;
                \item resolving entities and predicates in an automated and reliable way, without relying on any third-party resources.
     		\end{itemize}
	\item we propose a ``zero-shot'' approach, as all the devised prompts do not need any examples or external KBs for inferring the related information.
	\item our proposal may deal with large-scale data, as no human effort and examples documents are required;
	%\item the pipeline integrates a novel strategy for merging the LLM outputs and performing entity/predicate resolution, i.e., to disambiguate the entities and relationships. In particular, the main innovation is a proposal of an automated approach that does not use any other external resources, such as knowledge bases or domain expertise;
	\item we exploit the outcome of several prompts for manually building a ground truth, useful to apply additional evaluation metrics.
\end{itemize}

\section{Methodology}
\label{sec:methodology}
This Section describes our methodology, starting from the general perspective, introducing the main key points and functionalities, and, subsequently, detailing the pipeline step by step.

\subsection{LLM Prompting}
\label{sec:llm_prompting}
The main focus of our innovative approach is the integration of task-specific prompting of an LLM in each step of the KG construction. To this end, we first investigate, in this work, the capability to analyze, elaborate, and generate human-like text of a trendy, well-known LLM, i.e., \gptmodel{}, in particular the \gptmodelrelease{} release\footnote{\url{https://platform.openai.com/docs/models/gpt-3-5}}. We access the \gptmodel{} capabilities using the official chat completion API\footnote{\url{https://platform.openai.com/docs/api-reference/chat/create}}. As described in the API references, we interact with the model submitting the chat as a list of messages, each with a specified role, and we get a response coherent with the conversation thread. There are four supported roles, but we are only focused on the following:

\begin{itemize}
    \item \textit{system}: the role which aims to tell the model how to behave in its responses;
    \item \textit{user}: the role of the user messages/requests;
    \item \textit{assistant}: the role of the model responses.
\end{itemize}

In our case, we pass the detailed task instructions as a system prompt and the data to operate on as an appropriately formatted user prompt. We receive the task results in a message with the assistant role. Furthermore, we aim to obtain a deterministic behavior of the model (i.e., to get the same message in response to the same input prompts) by setting the API \textit{temperature} parameter to zero.

\subsection{Methodology Overview}
\label{sec:overview}
\begin{figure}[h]
    \centering
    \includegraphics[width=0.9\textwidth]{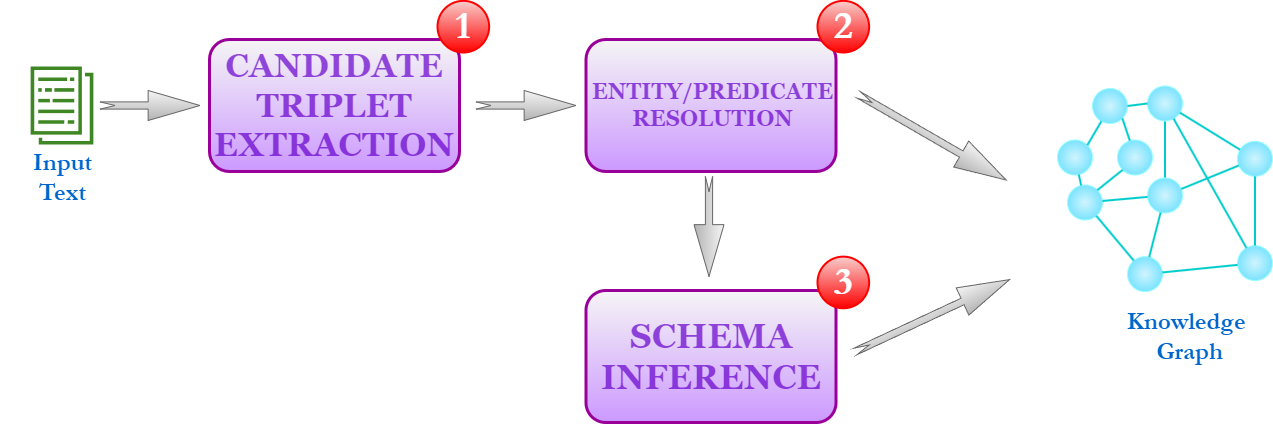}
    \caption{High-level architecture}
    \label{fig:overview}
\end{figure}

The main process of the proposed approach can be represented by the high-level architecture in Figure~\ref{fig:overview}, which describes the main stages of the process, and embodies three main tasks: \textit{\nameref{sec:overview_prompting}} (1), \textit{\nameref{sec:overview_resolution}} (2), and \textit{\nameref{sec:overview_inference}} (3). Such tasks are performed sequentially, and their main functionalities are described in the following sections.

\subsubsection{Candidate Triplet Extraction}
\label{sec:overview_prompting}
As already remarked, LLM prompting can leverage the capability of a model to process a massive amount of information. In this module, the challenge is defining well-formed prompts to effectively generate proper candidate triplets for being included in the final graph. In particular, we first established the following goals:

\begin{itemize}
	\item \textbf{A proper entity characterization}, aiming to explore further the simple identification of text spans representing an entity \textit{mention}, enriching a potential mention with:
	\begin{itemize}
		\item a representative \textit{entity label}, not necessarily corresponding to an exact text span;
		\item a proper entity \textit{description};
		\item a list of \textit{types} or \textit{hypernyms} that denote the entity beyond the mention.
	\end{itemize}
	\item \textbf{An appropriate characterization of triplets and predicates}, representing a relation between two entities (i.e., the \textit{subjet} and the \textit{object}) with a relevant predicate defined by:
	\begin{itemize}
		\item a suitable label, not necessarily corresponding to an exact text span;
		\item a general description of the relationship existing between subject and object.
	\end{itemize}
\end{itemize}

The requirement of an extended entity/predicate description comes out from providing a more informative and exhaustive representation of the extracted concepts, enriching the pure extraction approaches, like classical OpenIE methods, that are typically based on identifying and extracting entities or predicates corresponding to exact text segments included in the given input document. We deem that the description generated by our approach, together with the entity type list, provides a detailed context for each mention, leading us to an actual \textit{semantification}.
%

%\sgtcm{Questa parte a seguire dove si parla di LLM andrebbe spostata fuori da \nameref{sec:overview_prompting} perché ora si è previsto l'uso di LLM in tutte le fasi.}

%To this end, we first investigate, in this work, the capability to analyze, elaborate, and generate human-like text of a trendy, well-known LLM, i.e., \gptmodel{}, in particular the \gptmodelrelease{} release\footnote{\url{https://platform.openai.com/docs/models/gpt-3-5}}. We access the \gptmodel{} capabilities using the official chat completion API\footnote{\url{https://platform.openai.com/docs/api-reference/chat/create}}. As described in the API references, we interact with the model submitting the chat as a list of messages, each with a specified role, and we get a response coherent with the conversation thread. There are four supported roles, but we are only interested in three:
%\begin{itemize}
%    \item \textit{system} is the role to use to tell the model how to behave in its responses.
%    \item \textit{user} is the role of the user messages/requests.
%    \item \textit{assistant} is the role of the model responses.
%\end{itemize}
%In our case, we pass the detailed task instructions as a system prompt and the data to operate on as an appropriately formatted user prompt. We receive the task results in a message with the assistant role.

\subsubsection{Entity/Predicate Resolution}
\label{sec:overview_resolution}
On the one hand, the goal of the entity resolution is the identification and merging of mentions which refers to the same underlying entity. On the other hand, predicate resolution refers to identifying and resolving different expressions or textual forms that convey the identical relation between entities. As the data sources are mainly unstructured nowadays, one of the main challenges in this task is addressing the ambiguity and heterogeneity of natural language expressions without relying on external resources or domain expertise. Furthermore, capturing the complete semantic knowledge of entities or predicates is still complicated, particularly when dealing with domain-specific or complex concepts. To this end, we devised a novel resolution model which combines a semantic aggregation of concepts with a subsequent prompting stage. The former aims to identify and aggregate similar semantic concepts, and the latter aims to identify, in such groups, the concepts having the same meaning and labeling with a unique textual representation. A preliminary clustering is necessary as the current LLMs are more efficient with limited information. Hence, sending all the extracted entities or relations may be less reliable. The method is detailed in Section~\ref{sec:resolution}.

\subsubsection{Schema Inference}
\label{sec:overview_inference}
A KG schema describes the design, organization, and semantics of a given KG. In detail, it defines the types of entities, including their attributes and relationships, acting as a blueprint or a conceptual model for a KG. Providing a KG with an appropriate schema facilitates the reuse, sharing, and comprehension of the graph for humans and machines.
Nevertheless, building a schema from scratch is challenging, as, typically, it is often an interactive task that requires domain expertise and feedback by human knowledge. We aim to infer a schema in an automated way without relying on any human support. To this end, we deem that LLM prompting can be helpful also in this stage.

\subsection{Knowledge Graph Construction Pipeline}
\label{sec:pipeline}
\begin{figure}[h]
    \centering
    \includegraphics[width=0.7\textwidth]{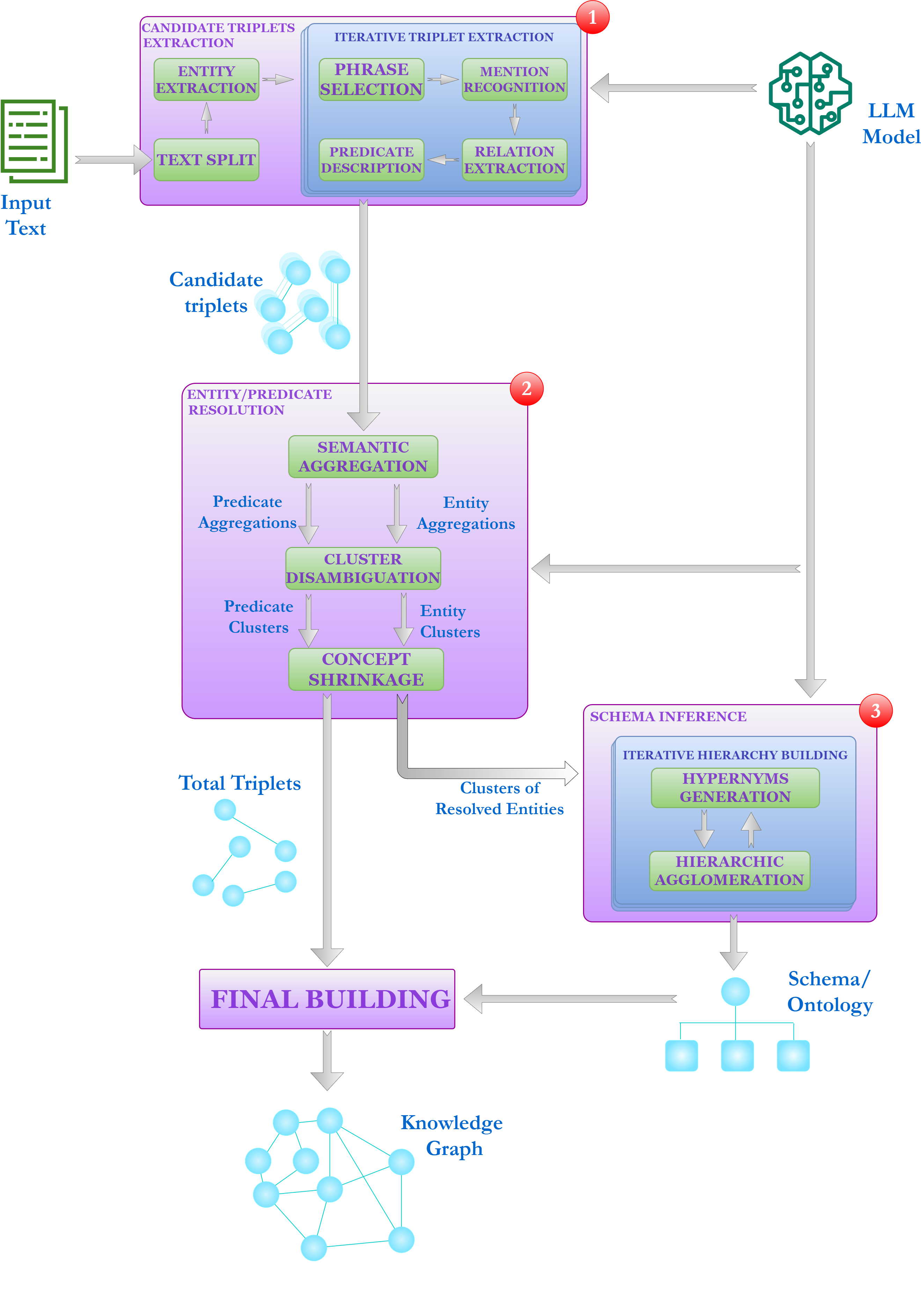}
    \caption{KG generation pipeline}
    \label{fig:pipeline}
\end{figure}

The devised KG generation pipeline is detailed in Figure~\ref{fig:pipeline}. The Figure provides an in-depth summary of all modules mentioned in the previous Section, described in the following.

\subsubsection{Candidate Triplet Extraction}
\label{sec:prompts}

Defining the LLM-prompting approach for extracting well-characterized entities and triplets requires decomposing the problem into simpler tasks resolvable by \gptmodel{} and studying the most effective prompts to solve them. To this end, we performed an exploratory stage to find an optimal prompting strategy: the task decomposition and prompt definition activities have been iterated in a trial-and-error method until each task performed with adequate reliability and accuracy. We started by using the official guidelines\footnote{\url{https://platform.openai.com/docs/guides/gpt-best-practices}} as a reference and gradually leveraged the experience gained during the various trials to finally define the current candidate triplet extraction phase.

%\begin{figure}[h]
%    \centering
%    \includegraphics[width=0.9\textwidth]{prompting.png}
%    \caption{Extraction by prompting pipeline}
%    \label{fig:prompting}
%\end{figure}

% {\textcolor{teal}{SGT - Stavo pensando che sposterei \textit{text split} come passo di pre-procsseing fuori dall'\textit{LLM prompting} (ovvero nella pipeline globale) spostando il testo di \textit{text split} come sotto-sezione nell'overview, e qui così parlo solo di estrazione. Tuttavia dato che nel \textit{text split} è prevista la summarization potrebbe avere molto senso farlo ricadere in \textit{LLM prompting}.}}

The devised triplet extraction approach begins with a pre-processing stage (\textit{text split}) designed to reduce overly long texts into smaller chunks to process for the extraction by \gptmodel{}. Inspired by what might be an intuitive and systematic way of approaching the same extraction task by a human annotator, the subsequent extraction stage from the text chunks decomposes into two phases, in which we first identify the entities in a text chunk (\textit{entity extraction}) and then check how they relate to each other to find the actual triplets (\textit{iterative triplet extraction}). We deem that this is a more natural and straightforward process than directly starting with the triplet extraction, in which the search for entities is implicit and adds to the complexity of identifying the relationships, also simplifying the overall task and dealing with the difficulty of \gptmodel{} in reliably performing long and complex operations. The iterative strategy addresses the additional struggle of \gptmodel{} in respecting some constraints, which in this case means explicitly referencing the original list of entities when searching for triplets.

We cover the details of the presented problem decomposition in the following sections.

\newpage

\textbf{Text Split} 

\noindent Generative LLMs have a limit on the number of tokens they can process due to architectural and computational constraints. The specific \gptmodel{} has a limit of 4,096 tokens (defined by the cl100k tokenizer\footnote{\url{https://github.com/openai/openai-cookbook/blob/main/examples/How_to_count_tokens_with_tiktoken.ipynb}}), shared by both the received input and the generated output. Because of this technical limit, entity, and triplet extraction from long texts\footnote{We refer to long texts that are not decomposable into several stand-alone text excerpts but consist of a single unique narrative thread.} must be performed in several steps to ensure the token limit is never reached.

Splitting the text into several independent chunks is a non-trivial necessity that brings to addressing two major problems:

\begin{itemize}
    \item The \textbf{omission of the global context}, since each independent chunk use terms and expressions that may lose their meaning when extrapolated from the overall narrative of the full text.
    \item The \textbf{separation of related entities}, as two entities that are members of a triplet may fall into two separate chunks, making it impossible to recover their relationship.
\end{itemize}

We tackled the two problems by designing a splitting technique that aims to:

\begin{itemize}
    \item \textbf{Avoid the loss of context} by providing each chunk with a summary of the preceding chunks to serve as a global context. Since the summarization task also faces the token limits, the summary for $chunk_{i}$ (i.e the summary of all chunks from $1$ to $i-1$ and that we call $summary_{i}$) is generated using $chunk_{i-1}$ and $summary_{i-1}$ as input, as formally stated in the formula:
    \[summary_{i} = summarization\_task(summary_{i-1}, chunk_{i-1})\]
    Since we process the chunks in order when we generate $summary_{i}$ we already have both of the necessary inputs as they have been computed with the extraction on $chunk_{i-1}$.
    \item \textbf{Reduce the probability of separating two related entities} by defining chunks as partially overlapping sliding windows of the input text: the larger the chunk size and overlap, the lower the probability of separating the two related entities.
\end{itemize}

\textbf{Entity Extraction}

The extraction task begins by identifying entity mentions in a text chunk. On the one hand, we guided \gptmodel{} using a detailed system prompt that respects this overall specifics:

%\begin{enumerate}
 %   \item \label{item:ee_entity_detail} An explanation of what is meant by an entity;
  %  \item \label{item:ee_instruct} An explicit request to retrieve the entity mentions in the user's text, specifying the additional requirement of providing a description and a list of types for each;
   % \item \label{item:ee_format} The details for a properly formatted output.
%\end{enumerate}

\begin{itemize}
    \item[S1)] label{item:}An explanation of what is meant by an entity;
    \item[S2)] An explicit request to retrieve the entity mentions in the user's text, specifying the additional requirement of providing a description and a list of types for each;
    \item[S3)] The details for a properly formatted output.
\end{itemize}

The text from which to extract the entities is integrated within the user's prompt.

While S2 explains what task will be performed and S3 establishes a standard format for the output, the actual role and functionality of S1 are less immediate. S1 is not mandatory for the actual execution of the task since \gptmodel{} already \qt{knows} what an entity is in the context of knowledge graphs and can find entity mentions even without our specifics. However, there is a notable difference in the amount and significance of the results extracted when we explicitly provide \gptmodel{} with our description of what we mean by an entity. Including S1 decreases the number of entities retrieved but leads to a greater focus on concrete nouns and named entities better detailed in the text chunk, leaving general and abstract nouns to serve mainly as types. Moreover, when we exclude S1 from the prompt, we observe less detailed entity descriptions, which we can especially appreciate by comparing the entities commonly found with and without S1.

For example, we refer to a few entities extracted from the use case texts introduced in Section~\ref{sec:experiments}. From the textual content
of \qt{Cagliari} webpage, we extract 27 entities with respecting the S1 constraint and 62 when excluding it. However, the extra entities are mostly general concepts such as \qt{History and art}, \qt{Sea and parks}, \qt{Sea view}, \qt{Shopping streets}, and many others, which are not well covered in the text. Generating these entities has a negative effect on the descriptions themselves, which cannot be detailed. Furthermore, this negative impact also affects the entities that are properly described: thus, the description of \qt{Cagliari} passes from \qt{The capital city of Sardinia, offering history, art, seashores, parks, and fine cuisine} with S1 to \qt{The capital of Sardinia}" without S1; the description of \qt{Marina District} changes from \qt{A quarter in Cagliari, featuring lovely buildings and the porticos of Via Roma, including the Palazzo Civico} to \qt{The waterfront district of Cagliari}.

Because of this observed behavior, we evaluated the contribution of S1 as positive and maintained it in our implementation and experiments. We denote as \es{} the set of entities extracted at this phase.

\textit{Iterative Triplet Extraction}.
\label{sec:prompts_triplet_extraction}
In this phase, we use the set of entities \es{} as a constraint to extract the relations between the entities from the text chunk content \tf{} in the form of triplets. Extracting triplets that explicitly refer to the set of entities \es{} poses a challenging requirement for \gptmodel{}. When set \es{} is highly populated, \gptmodel{} fails at sticking to the complete set and starts generating triplets involving entities not in \es{}. Even if some of these triplets can be considered an expression of correct statements, they are certainly not very interesting since they include entities for which we don't have a description or a list of types, which are an essential requirement of our approach.

%\sgtcm{da qui ...}

%We considered the possibility of collecting the new entities to later ask \gptmodel{} for a description and list of types, but we found it either:
%\begin{enumerate}
%    \item \label{item:difficult} Difficult, since we would ask again to perform a not straightforward task respecting a close set E to reference, which is the same problem that brought us into facing this problem in the first place.
%    \item \label{item:not_smart} Not very smart since we would perform the entity extraction phase knowing in advance that we would probably have to repeat a part of it on the new non-detailed entities.
%\end{enumerate}
%\Cref{item:not_smart} also led us to the logical question: why don't we start with the triplet extraction and then ask \gptmodel{} to detail all the extracted entities later? We conducted some experiments to explore this alternative, but we found it easier to tackle the problem posed by \cref{item:difficult} when starting with the generation of the entities rather than the triplets.

%\sgtcm{... fino a qui si può anche tagliare. Si tratta di una digressione con qualche dettaglio e considerazione, ma se dovesse servire spazio secondo me è eliminabile facilmente e senza rimorsi.}\textcolor{red}{ALE: o al limite spostarla nella parte di discussione esperimenti, se non vogliamo buttarla via.}

To extract triplets that refer only to entities in \es{}, we designed an iterative process in which we extract triplets by focusing on a different entity \ei{} $\in$ \es{} at each iteration $i$. The core concept of each iteration is to simplify the task complexity at every step, making it easier to select only entities in \es{}. We discuss the steps in each iteration in the following paragraphs.

%To extract triplets that refer only to entities in \es{}, we designed an iterative process in which we extract triplets by focusing on a different entity \ei{} $\in$ \es{} at each iteration. We discuss the steps of this process in the following paragraphs.

\textbf{Phrase Selection}

To focus on \ei{} and avoid using the whole text \tf{}, which might contain unrelated statements only about other entities, we aim to extract a target text excerpt \te{} for the triplet extraction that summarizes the information about \ei{} present in \tf{}. \te{} defines a smaller context than the complete \tf{}, thus reducing the complexity of the next extraction steps and increasing the reliability of \gptmodel{}.

The extraction of \te{} may be compliant with classical \textit{Query-Focused Abstractive Summarization} or \textit{Open-Domain Question Answering} tasks that can be performed again with \gptmodel{}. We designed a  suitable prompt focusing on the abstraction of simple declarative sentences that provide information about the entity \ei{}, which we can concatenate to obtain \te{}. We also identified an alternative solution in which we bypass the explicit extraction of \te{} and directly select the very same description of \ei{} already generated with the approach described in Section \textit{Entity Extraction}. The latter solution provides a more concise (but still meaningful) \te{} than the one obtainable with the former solution but significantly reduces the computation time and cost due to the omitted API calls. The experiment in \Cref{sec:experiments} uses the latter solution.

\textbf{Mention Recognition}

The new \te{} text is a summary focusing on \ei{}, in which we expect to find mentions of the other entities \ej{}$\in$\es{} that are related to \ei{} and contribute to shaping its identity. We want to detect which entities \ej{} are mentioned in the generated summary, thus defining the subset \esi{} of entities mentioned in \te{}. We expect \esi{} to have significantly fewer entities than \es{}.
To solve this task, we instruct \gptmodel{} with a system prompt requesting to recognize the mentions of the listed entities in the specific text, all provided by the user. We ask \gptmodel{} to rewrite the same list of entities, appending a \qt{yes/no} answer at the end of each entry. The user prompt includes the numbered list of all the entities in \es{} and the generated text \te{}, appropriately delimited and marked.

Although we demand to refer to the complete set of entities \es{}, we enforce this requirement on a simple task to perform on a small reduced-context \te{}, which allows \gptmodel{} to do it reliably. Moreover, using a numbered list instead of a simple bulleted list introduces the necessity to preserve the association entity ID/label, which seems to support \gptmodel{} in a correct execution even more. We retrieve the mentioned entities \ej{} that define \esi{} by finding the entries marked with a \qt{yes} in the answer.

\textbf{Relation Extraction}

We can perform the relation extraction using the narrowed context constructed in the previous two steps. We query \gptmodel{} with the system prompt for identifying the relations between the listed entities within the text, both supplied by the user. We ask to express the entity relations in the form of RDF triplets, using subjects and objects selected from the list of entities (reporting both their name and ID) and by choosing an expressive predicate. The user prompt provides the text \te{} and the numbered list of entities in \esi{}, appropriately delimited and tagged. \gptmodel{} answers with the identified triplets \re{}.

The system prompt includes an essential clarification of what we mean by an expressive predicate, improving the overall quality of the extracted triplets. Our explanation points \gptmodel{} into generating predicates that correctly represent the relationship between the two entities without being too specific, as it would make the predicate hardly reusable and observable in other triplets, aiming for a sort of predicate \textit{canonicalization}.

We refer to an extracted triplet to better explain the importance of our addition. Given the following text excerpt:

\begin{quote}
\textit{\textbf{Cagliari}: the capital of Sardinia is steeped in Mediterranean atmosphere and offers everything you could want from a vacation: history and art, seashores and parks, comfort and fine cuisine. Picturesque historical districts with sea views, elegant shopping streets and panoramic terraces, including the \textbf{Bastione di Santa Croce}, a great place for a romantic evening after a fiery sunset.}
\end{quote}

Among the entities found with the entity extraction prompts we can find:

\begin{itemize}
    \item \textbf{Cagliari} (City, Tourist Destination): The capital city of Sardinia, offering history, art, seashores, parks, and fine cuisine.
    \item \textbf{Bastione di Santa Croce} (Tourist Attraction, Landmark): A panoramic terrace in \textbf{Cagliari}, offering a romantic view of the sunset.
\end{itemize}

With which we extract the triplet:

\begin{itemize}
    \item \textbf{Cagliari}; \textit{has landmark}; \textbf{Bastione di Santa Croce}
\end{itemize}

According to the provided text, it could be extracted other overly specific predicates like \qt{has panoramic terrace} or \qt{has great place for romantic evening}, or other excessively generic ones like \qt{includes}. We did not choose these predicates randomly but used explicit terms within the text as most OpenIE tools would do. We deem the choice of a predicate like \qt{has landmark} to be much more meaningful, as it better expresses the inferable relationship between those two entities. We observed the same tendency in the choice of the predicate in the other extracted triplets.

Since both \te{} and \esi{} are reasonably simple, \gptmodel{} has proven capable of reliably extracting meaningful triplets while respecting the given list of entities, unlike the naive case where we refer to the complete text \tf{} and the full set of entities \es{}.

\textbf{Predicate Description}

To avoid adding complexity to the relation extraction step, we exclude the generation of the predicate description, which we set as one of our main goals, choosing to do it in a final independent step. We use a system prompt to tell \gptmodel{} to return the description of each unique predicate, referencing the text and list of RDF triplets provided by the user. The user prompt provides the text \te{} and the list of triplets \re{}, appropriately delimited and tagged. The model responds with a list of predicate and description pairs.

The complexity that prevented us from generating the description during the relation extraction arises from the need to have not just any description but one capable of capturing the generic nature of the relation expressed by the label of the predicate. We defined this necessity in our system prompt. We refer again to the extracted triplet used earlier as an example:

\begin{itemize}
    \item \textbf{Cagliari}; \textit{has landmark}; \textbf{Bastione di Santa Croce}
\end{itemize}

Simply asking for a predicate description leads to answers like \qt{It expresses that the Bastione di Santa Croce is a landmark in Cagliari}, which are tightly related to the specific instance of the relation in which we use that predicate. With our prompt instead, we can get a description like \qt{Expresses a relationship between a place and a landmark located in it}, which perfectly fits the predicate and our original intentions. We obtain the same behavior during the description of most of the predicates.

\textit{Response validation.}
Each system prompt specifies to \gptmodel{} the format to respect when generating the response, which is necessary to parse the response and retrieve the results. For the tasks where we expect a list of results, the format details are for a generic line of the list. For the other tasks involving numbered lists of entries to refer to, we designed the output format to enforce the model into reporting both the entry label and number, which seems to increase the reliability of the answers.

According to these prompt standards, we introduce two types of tests on the model answers:
\begin{itemize}
    \item \textbf{Pattern matching} (always) uses regular expressions to check that the answer, or each of its lines, correctly conforms to the format described in the system prompt.
    \item \textbf{Consistency check} (only when we reference a numbered list) verifies that the model preserves the label/ID association, checking  if that label corresponds to that numbered entry.
\end{itemize}

When a response (or one of its lines) doesn't pass all the required tests, we discard it as unparsable or likely to contain false information.

\subsubsection{Entity/Predicate Resolution}
\label{sec:resolution}
As already mentioned, the module combines semantic aggregation with proper LLM prompting. Semantic aggregation aims to identify and aggregate related entities/relations in groups containing concepts having similar semantic meanings or referring to a higher-level concept (e.g., ``car'' and ``motorcycle'' are related to the concept of a ``vehicle''). This preliminary aggregation is needed to split the information and send a sequence of prompts rather than a unique prompt. Indeed, as also reported in the GPT models guidelines\footnote{\url{https://platform.openai.com/docs/guides/gpt-best-practices/six-strategies-for-getting-better-results}}, complex tasks tend to have higher error rates than simpler tasks. The entity/resolution task is detailed in the following:

\textbf{Semantic aggregation}

The first step is to aggregate all semantically similar entities and do the same with the relations. In detail, we devised a proper aggregation strategy, based on computing two specific similarity scores for entities ($S_{e}$) and relations ($S_{r}$), both based on analyzing the contributions of the KG components (entity/relation, description, and type). To this end, we first consider all the pairs combinations for both entities and relations. For each pair ($i, j$) of entities or relations, we compute the individual contributions as follows:

\begin{itemize}
	\item \textit{Label similarity}. We computed a similarity score for each entity or relation pair $(i, j)$ as the Levenshtein distance \cite{bib:levenshtein66} between the two entity labels. We denote as $e_{i,j}$ the similarity between two entities, and with $r_{i,j}$ the similarity between two relations. Then, $e_{i,j}$ and  $r_{i,j}$ are normalized in the range $[0,1]$, where a similarity equal to 1 means the entities are identical.
	\item \textit{Entity types similarity}. We adopt the same strategy of entity and relation similarity for computing the similarity ($t_{i,j}$) between each type pair $(i, j)$. Let us remark that types are associated with entities only.
	\item \textit{Description similarity}. As the description is a text segment more complex than a simple entity or type label, we project all entity and relation descriptions in an embedding space, and we compute the similarity between two descriptions $i$ and $j$, relying on a classical cosine similarity metric. In detail, we adopt the Universal Sentence Encoder model \cite{bib:cer18} as the embedding model, and we denote as $ed_{i,j}$ the similarity between two entity descriptions, and with $rd_{i,j}$ the similarity between two relation descriptions.
\end{itemize}
The final similarities scores are, in both cases, a weighted combination of the previous contributions and are summarized by the following formulas:

\begin{equation}
	S_{e}(i,j)= \alpha \cdot e_{i,j} + \beta \cdot ed_{i,j} 
\end{equation}

\begin{equation}
	S_{r}(i,j)= \gamma \cdot e_{i,j} + \delta \cdot rd_{i,j} \\
\end{equation}

\noindent We empirically fixed the coefficient values, choosing $\alpha=0.35$, $\beta=0.65$, $\gamma=0.25$, and $\delta=0.75$. 

The final goal of this task is to aggregate similar entities/relations. In so doing, we adopted distinct empiric strategies for entities and predicates.

\textit{Similar entities}: two entities $i$ and $j$ are considered similar if:
\begin{itemize}
	\item $S_{e}(i,j) \geq 0.9 \; \; \;$ or
	\item $0.7 < S_{e}(i,j) < 0.9 \;$ and $\; t_{i,j} > 0.25$. 
\end{itemize}

The latter condition is to give importance to a pair if they are less similar but have somewhat related types.

\textit{Similar predicates}: two relations $i$ and $j$ are considered similar if:
\begin{itemize}
	\item $S_{r}(i,j) \geq 0.8$
\end{itemize}

The module outputs a set of aggregations for both entities and relations. Each aggregation is a group of semantically similar entities or relations. In other words, each group represents a \textit{cluster} of entities/relations. Each cluster is integrated into a proper prompt sent to the LLM. As mentioned above, splitting the elements into clusters is needed to improve the performance of LLM prompting.

\textbf{Cluster disambiguation}

Each entity/relation cluster contains semantically similar elements, but typically they are not all related to only one concept; it is actually what we expect from semantic aggregation. As an example, let us suppose a cluster composed of the entities \textit{car}, \textit{automobile}, \textit{motorcycle}, \textit{motorbike}, \textit{bicycle}, and \textit{bike}. They are similar, as all entities are means of transport but do not represent the same entities. The goal of this task is to identify which subsets refer to the same entity; in the previous example, \textit{car} and \textit{automobile} refer to the same entity, like \textit{motorcycle} and \textit{motorbike}, and \textit{bicycle} and \textit{bike}.

To this end, we prompt the GPT-3.5 model with another well-formed prompt, asking to return the subsets of semantically equal entities or relations. Iterating across all the clusters, the outcome of this task will be a set of semantically identical entities or relations.

\textbf{Concept shrinkage}

Let us remark that entity resolution aims to identify entity mentions that refer to the same concept, providing a unique entity identifier for the aforementioned mentions. Likewise, predicate resolution recognizes relations having a different textual representation but conveying an identical relationship, denoting such relations with a unique relation identifier. To this end, each group of equal entities or relations returned by the \textit{Cluster Disambiguation} module is used to compose a further prompt aimed at asking for a unique label for representing the underlying group. Such a label is used as the final unique representation of the given entity/relation. 

\subsubsection{Schema Inference}
\label{sec:inference}
A schema is an underlying structure of a KG, usually a taxonomy or an ontology that accurately reflects the KG content. The schema contains the entity types, usually organized in a taxonomic or ontological structure, connected by proper relations. A suitable schema may be helpful in discovering implicit knowledge and supporting the exploitation of KGs in many tasks, e.g., in question answering or recommendation. Typically, a schema is manually defined by human experts, requiring considerable effort and time. To overcome this limitation, we aim to develop an iteratively bottom-up LLM-based schema inference. 

The proposed method is performed by module 3 (Schema Inference), depicted in Figure~\ref{fig:pipeline}. The module receives the set of clusters composed of resolved entities from which the corresponding types are considered. The schema inference is an iterative process that involves two main steps at each iteration:

\textbf{Hypernym Generation}

For each cluster, after removing the possible duplicates (as, obviously, different entities may belong to the same type), the types are embedded in an appropriate prompt sent to \gptmodel{} to find a common hypernym for the entire cluster and relation that links such hypernym to the entity types, or, depending on the cluster size and semantic similarities among types, finding a set of appropriate hypernyms, each one being related to a distinct cluster subset. For example, for the types \textit{legumes}, \textit{green vegetables}, \textit{poultry}, \textit{pork}, \textit{fish}, and \textit{crustacean}, the most suitable hypernyms may be \textit{vegetables} (connected to the types \textit{legumes} and \textit{green vegetables}),
\textit{meat} (for \textit{pork} and \textit{poultry}), and \textit{seafood} (for \textit{fish} and \textit{crustacean}). In all three cases, each type may be linked with the related hypernym with the relation \textit{is type of}. 

\textbf{Hierarchical Agglomeration}

Subsequently, all generated hypernyms and relations are merged across all clusters to remove redundancies. Let us point out that the initial entity types will represent the lower level of the schema, whereas the hypernyms will represent the upper level of the taxonomy. Afterward, for the upper level, we apply the same \textit{Semantic Aggregation} technique described in Section~\ref{sec:resolution} for finding a new set of clusters. We then applied the \textbf{Hypernym Generation} and the \textbf{Hierarchical Agglomeration} iteratively for constructing the upper levels of the taxonomy until we reach the scenario in which, at this stage, only one cluster and one hypernym are generated.

\section{Experiments}
\label{sec:experiments}
%

% This section describes the experiments we performed to validate the \nameref{sec:overview_prompting} starting stage of our pipeline, in particular, the extraction capabilities of \gptmodel{}. To test its validity, we identified a set of texts of interest from which to extract the triplets, along with the evaluation method and metrics that could quantify the quality of the extraction results in all their aspects (labels, descriptions, and types).

This section describes the experiments we performed to validate our pipeline and the extraction capabilities of \gptmodel{}. To test its validity, we identified a set of texts of interest from which to extract the triplets, along with the evaluation method and metrics that could quantify the quality of the extraction results in all their aspects (labels, descriptions, and types).

\subsection{Dataset}
\label{sec:dataset}

As our input data, we gathered a set of informative texts from the English version of the SardegnaTurismo website\footnote{\url{https://www.sardegnaturismo.it/en/}}, which is a portal describing tourist destinations and points of interest in Sardinia. We targeted the city of Cagliari (the capital of Sardinia), selecting 44 pages describing the city and nearby locations and providing a set of independent texts on the same topic in which recurrent mentions of the same entities are inevitably present.

The selected texts have an average of $\sim$660 tokens, with a peak of $\sim$1100 tokens (referring to \gptmodel{}'s tokenizer: cl100k\footnote{\url{https://github.com/openai/openai-cookbook/blob/main/examples/How_to_count_tokens_with_tiktoken.ipynb}}). Since each independent text afforded us to stay very far from the model's token limit, we did not undergo the need to split the text before processing, which was therefore not tested in this experiments.

\subsection{Evaluation}
\label{sec:evaluation}
The evaluation of a KG is a challenging key topic. Assessing the quality of the defined entities and relationships is crucial for ensuring the relevance, reliability, and suitability of the information held in the graph. Furthermore, another topic is to guarantee the inclusion of all relevant information. We evaluate the quality of the generated entities and triplets, relying on a manual annotation, where human assessors judge each component of the resulting KG. From the annotations, we compute several metrics described in the following Sections. 

\subsubsection{Human assessment}
\label{sec:human_assess}
A classical approach for evaluating the quality of information held by an AI model involves human assessors in judging and annotating the output of a system. In our study, human expertise may be highly effective for estimating the quality and the correctness of a generated KG. Furthermore, they can also support us in estimating completeness, e.g., identifying missing entities that should be included in the graph. 

\textit{KG components annotation.}
We asked several assessors to judge as \emph{correct} or \emph{incorrect} the following components:

% \begin{itemize}
% 	\item \textit{Entity}, which we judge as correct depending on two factors: (i) the entity should be mentioned in the input text\sgt{, which refers to an entity corresponding to that label and description}, and (ii) the entity should be related to the given context;
% 	\item \sgtcm{Non c'è annotazione di questo, se la descrizione non corrisponde all'entità citata nel testo, è sbagliata proprio l'entità \sout{\textit{Entity description}, considered correct if it is representative of the given concept, otherwise is labeled as incorrect;}}
% 	\item \textit{Entity type}, which is considered correct if it captures the actual \sgt{class\sout{context}} of the entity;
% 	\item \textit{\sgt{Triplet\sout{Relation}}}, labeled as correct depending on two factors: (i) the linked entities should \sgt{be correct\sout{exist (i.e., it links two correct entities)}}, and (ii) it should \sgt{\sout{be}represent a true and }reasonable \sgt{relation accurately expressed by the predicate label and description}; \textcolor{blue}{(TODO: inserire esempi)}
% 	\item \sgtcm{Come prima, non c'è annotazione di questo, se la descrizione non corrisponde alla relazione che sussiste tra le entità, è sbagliata proprio la tripla. La descrizione dice cosa sia la natura della relazione, più che la label. \sout{\textit{Relation description}, assessed in the same way as entity descriptions.}}
% \end{itemize}

\begin{itemize}
    \item \textit{Entity}, which we judge as correct depending on two factors: (i) the input text should mention the entity, i.e. there should be an actual reference to an entity corresponding to that label and description, and (ii) the entity should be intuitively significant and relevant to the overall textual context. To illustrate these criteria, we report two entities that were evaluated as \emph{incorrect}::
    \begin{itemize}
        \item \textit{label}: Santo Stefano del Lazzaretto \\
              \textit{description}: A location near the Tower of Prezzemolo. \\
              \textit{evaluation}: \emph{incorrect}. The entity violates (i) because although there is a Santo Stefano del Lazzaretto in the text, it is not a location near the Tower of Prezzemolo but another name for the same Tower of Prezzemolo.
        \item \textit{label}: Nature \\
              \textit{description}: The natural environment in Monte Claro park, including trees, bushes, and plants. \\
              \textit{evaluation}: \emph{incorrect}. The entity violates (ii) because, although the label/description could be a suitable pair to refer to the natural amenities of the park, at least in the narrow context of the text from which it comes, the entity is overly abstract, and the text barely addresses this concept.
    \end{itemize}
    \item \textit{Entity type}, which is considered correct if it captures the actual class and context of the entity. We provide an example:
    \begin{itemize}
        \item \textit{label}: Gaetano Cima \\
              \textit{description}: Famous architect who designed the facade of San Giacomo church. \\
              \textit{types}: [Architect, Neoclassical architecture] \\
              \textit{evaluation}: [(Architect, \emph{correct}), (Neoclassical architecture, \emph{incorrect})]
    \end{itemize}
    \item \textit{Triplet}, labeled as correct depending on two factors: (i) the linked entities should be correct, and (ii) it should represent a true and reasonable relation accurately expressed by the predicate label and description. We include some clarifying examples:
    \begin{itemize}
        \item \textit{triplet}: Nature; is part of; Monte Claro \\
              \textit{predicate description}: Expresses a relationship of inclusion or belonging between two entities. \\
              \textit{evaluation}: \emph{incorrect}. The entity violates (i) because Nature is \emph{incorrect}, as we said in the previous example.
        \item \textit{triplet}: Artistic exhibition; is showcased in; Museo del Tesoro \\
              \textit{predicate description}: None. \\
              \textit{evaluation}: \emph{incorrect}. The entity violates (ii) because the predicate description is missing, so the predicate cannot represent any relation.
        \item \textit{triplet}: Casa Spadaccino; has garden; Outdoor activities \\
              \textit{predicate description}: Expresses the presence of a garden in a location. \\
              \textit{evaluation}: \emph{incorrect}. The entity violates (ii) as Outdoor activities is not a garden located in Casa Spadaccino and, indeed, it is not even a garden.
    \end{itemize}
\end{itemize}

\textit{Inferred components annotation.}
% Furthermore, both correct entity and relation descriptions are labeled with an additional annotation, i.e., if the entity/description has been generated using explicit or implicit information retrievable from the text or if the LLM generated the description inferring it from its inside knowledge.
Furthermore, both correct entities and triplets are labeled with an additional annotation to evaluate whether \gptmodel{} retrieves the information from the text or draws on its knowledge. For the entities, we check whether the description contains information from the given text or additional information generated by \gptmodel{}. For the triplets, we check whether the relation between the two entities that the predicate expresses can be inferred from the text or is just known a priori by the model.

As an example, the following entity is evaluated as having a \gptmodel{} generated description because there is no mention in the text of the global extent of the event or the start and end years of the war.

\begin{itemize}
    \item \textit{label}: WWII \\
          \textit{description}: An abbreviation for World War II, a global war that lasted from 1939 to 1945.
\end{itemize}

Instead, we found no triplets with a relation recognized by the model but missing from the text, and perhaps we cannot provide an example.

\textit{Ground truth annotation.}
The assessors provided a further annotation, identifying, for each document, a list of ``missed'' entities, i.e., relevant entities included in the input text but not retrieved by the model. Indeed, a challenge in listing the omitted entities in an open-domain setting is the ambiguity in defining which concepts should be of interest and thus extracted since there is no topic reference. Practically, anything could be considered an entity. Therefore, we decided to evaluate the coherence of the \gptmodel{} model in the entity extraction, using the types assigned to the entities to obtain a reference schema according to which it is possible to define which entities are missing. In detail, the assessors considered all entity types automatically extracted with at least two associated entities. For each type, they identified all entities mentioned in the input text that should have been labeled with the underlying type. Aggregating the missed and correct entities composes a suitable ground truth, useful for further evaluation of the final graph.

\subsubsection{Evaluation metrics}
\label{sec:metrics}
We rely on well-known metrics for assessing the extracted triplets. First, we adopt classical confusion matrix entries for assessing the performance in generating each component. In detail, each correct component is a \textit{true positive} ($TP$), an incorrect item is a \textit{false positive} ($FP$), and a missed entity is a \textit{false negative} ($FN$). Such entries permit us to compute the \textit{Precision} ($P$), i.e., the fraction of correct elements among all retrieved elements (see Eq.~\ref{eq:precision}), and the \textit{recall} ($R$), i.e., the fraction of correct retrieved elements among all correct elements (see Eq.~\ref{eq:recall}). In other words, precision estimates the ability to generate correct components, whereas recall evaluates the ability to identify all the relevant knowledge from documents. Furthermore, a useful metric that combines precision and recall is the \textit{F-score} ($F_{1}$), i.e., the harmonic mean of $P$ and $R$ (see Eq.~\ref{eq:fscore}).

\begin{equation}
	\label{eq:precision}
	P = \frac{TP}{TP + FP}
\end{equation}

\begin{equation}
	\label{eq:recall}
	R = \frac{TP}{TP + FN}
\end{equation}

\begin{equation}
	\label{eq:fscore}
	F_{1} = \frac{2 \cdot P \cdot R}{P + R}
\end{equation}

Let us note that the recall $R$ and $F_{1}$ can be computed only for entities, as, in this preliminary work, we asked assessors to annotate only the missing entities. Indeed, identifying information on missing relations is more challenging and requires more human effort. In detail, considering all generated components, we can compute the following metrics:

\begin{itemize}
	\item $P^{E}$: precision of the entity generation;
	\item $R^{E}$: recall of the entity generation; 
	\item $F^{E}_{1}$: F-score of the entity generation;
	\item $P^{T}$: precision of entity typing;
	\item $P^{R}$: precision of the relation extraction.
\end{itemize}

Furthermore, we can also estimate the ability of the model to infer additional information from its knowledge by taking into account, as already pointed out in the previous 
\Cref{sec:human_assess}, the correct entity descriptions and the relations which are not extracted from the input document. To this end, we define a score $\sigma$ corresponding to the percentage of truthful information returned by the \gptmodel{} model that comes from its internal knowledge ($I$) among all the returned truthful information ($D$):

\begin{equation}
	\label{eq:sigma}
	\sigma = \frac{I}{D} 
\end{equation}

Therefore, we can compute the sigma score with Eq.~\ref{eq:sigma} in these two variations:

\begin{itemize}
	\item $\sigma^{E}$: sigma-score of the entities description generation;
	\item $\sigma^{R}$: sigma-score of the relation extraction.
\end{itemize}

\subsection{Results}
\label{sec:results}

We applied our approach to the dataset described in \Cref{sec:dataset}, where each page represents an input document of the pipeline depicted in \Cref{fig:pipeline}. The system generated a basic knowledge graph comprising 761 entities and 616 triplets. Furthermore, the generated schema contains about $\approx 500$ nodes and $\approx 600$ edges. According to the assessment process described in the previous \Cref{sec:evaluation}, we manually annotated all the extracted graph components with a binary label to compute the evaluation metrics. We report the final results in \Cref{tab:computed_metrics}.

\begin{table}[h]
\centering
\begin{tabular}{@{}lccccccc@{}}
\toprule
Metric &
  $P^{E}$ &
  $R^{E}$ &
  $F^{E}_{1}$ &
  $P^{T}$ &
  $P^{R}$ &
  $\sigma^{E}$ &
  $\sigma^{R}$ \\ \midrule
Score (\%) &
  \multicolumn{1}{c}{98.82} &
  \multicolumn{1}{c}{93.18} &
  \multicolumn{1}{c}{95.92} &
  \multicolumn{1}{c}{85.71} &
  \multicolumn{1}{c}{75.31} &
  \multicolumn{1}{c}{ 9.20} &
  \multicolumn{1}{c}{ 0.00} \\ \bottomrule
\end{tabular}
\\[2mm]
\caption{Evaluation metrics scores.}
\label{tab:computed_metrics}
\end{table}

Let us point out that we have conducted experiments assessing only our approach since we cannot perform comparisons with state-of-the-art tools. The main reason is that there are yet no proven state-of-the-art tools to assume as a baseline, especially considering the peculiar kind of extraction performed by our method, which, to the best of our knowledge, is the only one that invests in detailing the entities with a description and a set of types and in providing a canonical label and extended description for the predicates used in the triplets. Furthermore, using custom corpora from a use case of our particular interest, chosen because of the general lack of datasets with sufficiently elaborate text adequate for testing our approach, prevented the possibility of retrieving the already available results from any other different tool that used the same dataset. We analyzed the results objectively, knowing that a comparison with the results from other toolkits will be required for fairness in the future, at least for the specific task we perform for which other toolkits are also available.

The results, considering the early stage of our study, are encouraging, as our approach and prompts guided \gptmodel{} in extracting correct and valid entities 98.82\% of the time ($P^{E}$). Furthermore, the system retrieved most of the entities within the same types, with a recall of 93.18\% ($R^{E}$) and an overall F-score of 95.92\% ($F^{E}_{1}$).

We also obtained a precision of 85.71\% in the typing task ($P^{T}$) and of 75.31\% in the triplet extraction ($P^{R}$), which we deem to be a very positive result given the open-domain setting, the zero-shot performance of the LLM model, and the complexity of the input text, as many of the correctly individuated relations are not blatantly explicit and involve entities that occur in distant clauses, as can be seen in the example shown in the \textit{Relation Extraction} paragraph of \Cref{sec:prompts_triplet_extraction}.

As expected, we obtained a higher precision in the task we considered easier and a slight precision decrease in the more challenging tasks of typing and relation extraction. The $\sigma$-scores tell us how often \gptmodel{} added its own knowledge when generating entity descriptions and relations. We observed the addition of new details in entity descriptions in 9.20\% of the cases ($\sigma^{E}$). However, since we extracted the triplets using the entity descriptions as contexts (see \textit{Phrase Selection} in \Cref{sec:prompts_triplet_extraction}), and we found no triplets describing relations that were not in the text ($\sigma^{R}$), this indicates that the details added in the entity descriptions are mostly marginal and do not constitute the whole sentence. The $\sigma$-scores also indicate another positive result, as we push the model to rely on what is in the text, helping to contain possible hallucination problems.

\section{Conclusion}
\label{sec:conclusion}

In this paper, we proposed a novel iterative approach to open-domain knowledge graph construction that leverages the zero-shot generative and comprehension capabilities of the trending \gptmodel{} model to address the typical challenges of this research topic caused by the lack of well-established datasets and highly reliable methods.

Our approach starts with the extraction of entities and triplets, providing the entities with multiple types and an extended description of their identity and the triplets with a proper explanation of the relation represented by the predicate and subsisting between the linked entities, going through an actual \qt{semantification} of the extracted elements. The candidate triplets are then refined in the following stages, exploiting the additional semantics offered by the types and descriptions both to formulate a suitable method to perform the entity and predicate resolution without relying on an already existing knowledge base and to infer a schema that defines the structure of the KG and simplifies its reuse and sharing.

Our experiments focused on evaluating the entities and triplets extraction capabilities of \gptmodel{} using web corpora with non-trivial content. The results show that \gptmodel{} is extremely good at entity extraction and performs very well in the typing and triplet extraction task, despite the open domain and zero-shot settings.

Although the lack of commonly available datasets makes it challengi-ng, we plan to conduct more in-depth experimentation of the proposed methods and properly compare our approach with other implementations on the same tasks. We also plan to explore the feasibility of introducing additional instructions in the prompts to define a scope, thus shifting the approach towards a closed-domain setting and focusing on the entities and triplets of particular interest. Finally, we will investigate the potential contribution of \gptmodel{}'s alternative generative LLMs, which are already emerging and will become increasingly popular in the future.

\section*{Acknowledgments}
 We acknowledge financial support under the National Recovery and Resilience Plan (NRRP), Mission 4 Component 2 Investment 1.5 - Call for tender No.3277 published on December 30, 2021 by the Italian Ministry of University and Research (MUR) funded by the European Union – NextGenerationEU. Project Code ECS0000038 – Project Title eINS Ecosystem of Innovation for Next Generation Sardinia – CUP F53C22000430001- Grant Assignment Decree No. 1056 adopted on June 23, 2022 by the Italian Ministry of University and Research (MUR).

 Furthermore, the authors thank Dr. Marco Manolo Manca, Diego Argiolas, and Gabriele Varchetta for their support in several activities of this work.

%Bibliography
\bibliographystyle{abbrvnat}  
\bibliography{references}  

\end{document}